\documentclass[letterpaper, 10 pt, conference]{ieeeconf}  % Comment this line out if you need a4paper

\IEEEoverridecommandlockouts                              % This command is only needed if
\usepackage{graphicx} % DO NOT CHANGE THIS
\usepackage{multirow}
\title{\LARGE \bf
Attention-Guided Lightweight Network for Real-Time Segmentation of Robotic Surgical Instruments
}

\author{%Zhen-Liang Ni$^{1,2}$, Gui-Bin Bian$^{1,2*}$, Zeng-Guang Hou$^{1,2,3}$,  \textit{Fellow, IEEE}, Xiao-Hu Zhou$^{2}$, Xiao-Liang Xie$^{2}$, Zhen Li$^{2}$% <-this %
Zhen-Liang Ni, Gui-Bin Bian$^{*}$, Zeng-Guang Hou, \textit{Fellow, IEEE}, Xiao-Hu Zhou, Xiao-Liang Xie, and Zhen Li% <-this %
\thanks{ Z. Ni, G. Bian, Z. Hou, X. Zhou, X. Xiao and Z. Li are with State Key Laboratory of Management and Control for Complex Systems, Institute of Automation, Chinese Academy of Sciences, Beijing 100190, China. Z. Ni, G. Bian and Z. Hou are also with the school of Artificial Intelligence, University of Chinese Academy of Sciences, Beijing 100049, China. Z. Hou are also with CAS Center for Excellence in Brain Science and Intelligence Technology, Beijing 100190, China. (Corresponding author: guibin.bian@ia.ac.cn)}
}
\begin{document}
\maketitle
%\thispagestyle{scrheadings} %forces first page to also have foot and header line
%\thispagestyle{empty} %forces first page to not have a header
%%\pagestyle{empty}
%
%\ieeefootline{Workshop on Latex Style Files \\ International Conference on Latex 2014, Las Vegas, NV, USA}%creates footline

%\ieeeheadline{Workshop on Latex Style Files \\ International Conference on Latex 2014, Las Vegas, NV, USA}%creates headline

%%%%%%%%%%%%%%%%%%%%%%%%%%%%%%%%%%%%%%%%%%%%%%%%%%%%%%%%%%%%%%%%%%%%%%%%%%%%%%%%
\begin{abstract}
The real-time segmentation of surgical instruments plays a crucial role in robot-assisted surgery. However, it is still a challenging task to implement deep learning models to do real-time segmentation for surgical instruments due to their high computational costs and slow inference speed.
In this paper, we propose an attention-guided lightweight network (LWANet), which can segment surgical instruments in real-time. LWANet adopts encoder-decoder architecture, where the encoder is the lightweight network MobileNetV2, and the decoder consists of depthwise separable convolution, attention fusion block, and transposed convolution. Depthwise separable convolution is used as the basic unit to construct the decoder, which can reduce the model size and computational costs. Attention fusion block captures global contexts and encodes semantic dependencies between channels to emphasize target regions, contributing to locating the surgical instrument. Transposed convolution is performed to upsample feature maps for acquiring refined edges. LWANet can segment surgical instruments in real-time while takes little computational costs. Based on 960$\times$544 inputs, its inference speed can reach 39 fps with only 3.39 GFLOPs. Also, it has a small model size and the number of parameters is only 2.06 M. The proposed network is evaluated on two datasets. It achieves state-of-the-art performance 94.10$\%$ mean IOU on Cata7 and obtains a new record on EndoVis 2017 with a 4.10$\%$ increase on mean IOU.

\end{abstract}
%However, real-time segmentation of surgical instruments using current deep learning models is still a challenging task due to the high computational costs and slow inference speed.
%However, real-time segmentation for surgical instruments remains a challenging task due to the high computational costs and slow inference speed of current deep learning models.
%However, real-time segmentation for surgical instruments using deep learning models is still a challenging task due to their slow inference speed.
%The lightweight network, MobileNetV2, is used as the encoder. Furthermore, a lightweight decoder is designed to recover the position details, which consists of attention fusion block, depth-wise separable convolution, and transported convolution. Also, specular reflection and class imbalance issues make this task more challenging. a deep separation convolution and attention fusion module. which consists of a lightweight network MobileNetV2 for extracting semantic features and a lightweight decoder for recovering positional details. The decoder consists of
%However, segmentation of surgical instruments using deep learning models cannot be applied to robot real-time control due to their slow inference speed.
%%%%%%%%%%%%%%%%%%%%%%%%%%%%%%%%%%%%%%%%%%%%%%%%%%%%%%%%%%%%%%%%%%%%%%%%%%%%%%%%
\section{INTRODUCTION}
In recent years, significant progress has been witnessed in robot-assisted surgery and computer-assisted surgery. Real-time semantic segmentation of surgical robotic instruments is one of the key technologies for surgical robot control. It can accurately locate robotic instruments and estimate their pose, which is crucial for surgical robot navigation~\cite{endovis2017}. Also, the segmentation results can be used to predict dangerous operation and reduce the risk of the surgery, contributing to achieving robotic autonomous operation. Furthermore, semantic segmentation of surgical instruments can provide a variety of automated solutions for post-operative work, such as objective skills assessment, surgical report generation, and surgical workflow optimization~\cite{Sarikaya,surgical2017ai,efpnet}. These applications can improve the safety of surgery and reduce the workload of doctors, which is significant for clinical work.
\begin{figure}[tbp]
  \centering
  \includegraphics[width=0.47\textwidth]{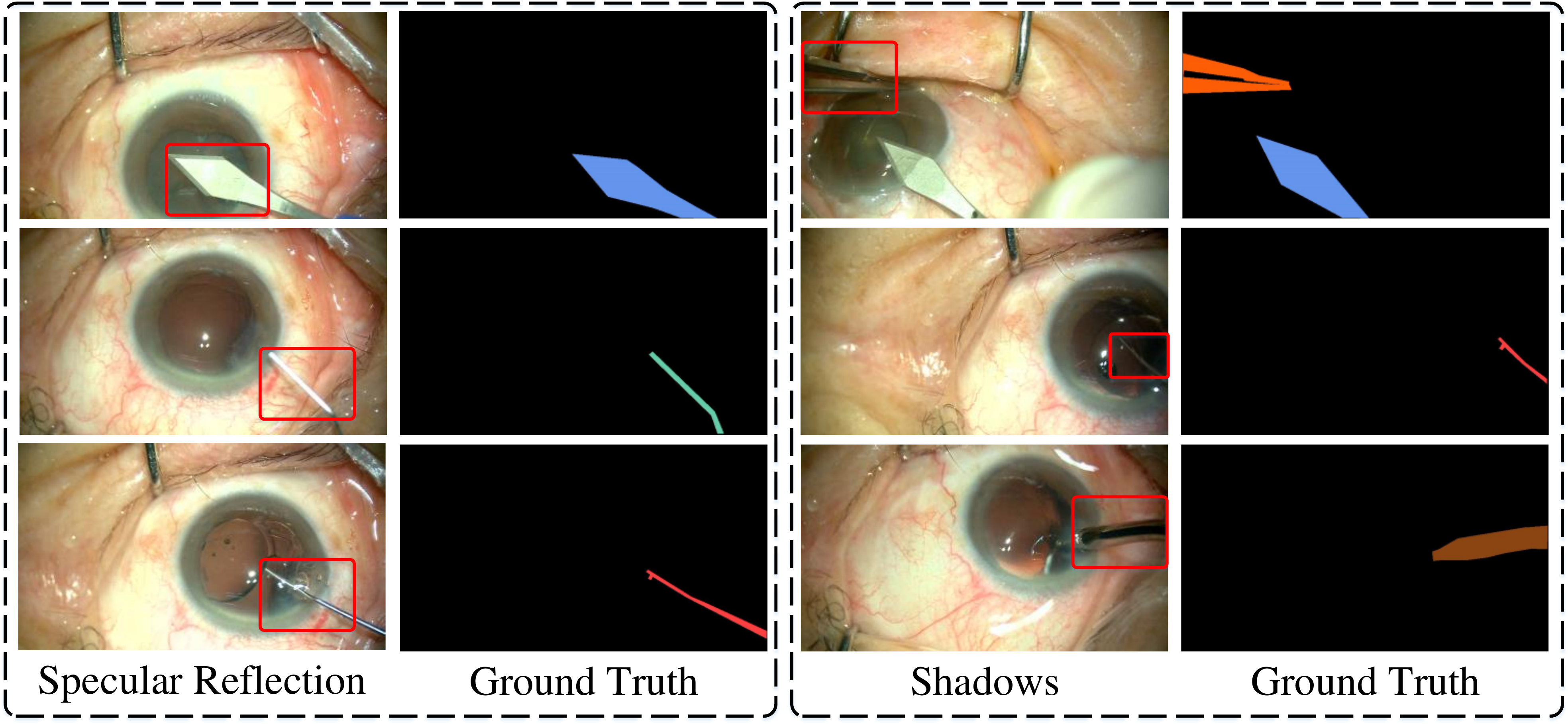}
  \caption{Challenges in semantic segmentation for surgical instruments. Different types of
  surgical instruments are marked by different colors. }
  \label{show}
\end{figure}

Recently, a series of methods have been proposed for the semantic segmentation of surgical instruments. The hybrid CNN-RNN method~\cite{attia} introduced Recurrent Neural Network to capture global contexts and expand receptive fields. RASNet~\cite{rasnet} adopted an attention mechanism to emphasize the target regions and improve the feature representation. Another work~\cite{qin} fused convolutional neural network prediction and the kinematic pose information to improve segmentation accuracy. However, those work mainly focused on fusing different forms of information for higher segmentation accuracy while failed to consider the inference speed, limiting their applications in real-time control of surgical robots.
%A work~\cite{luis} introduced optical flow to handle the deformation of surgical instruments.
%Due to limitations in computing resources, real-time segmentation of surgical instruments is still a huge challenge. Some existing networks can achieve good segmentation accuracy, such as TernausNet~\cite{ternausnet}, CSL Network~\cite{csl}. But these networks process images slowly, which cannot be used for real-time navigation of surgical robots. To address this problem, a new lightweight network is designed to achieve real-time segmentation of surgical instruments. There are some existing lightweight networks, such as MobileNet~\cite{mobilenet}, MobileNetV2~\cite{mobilenetv2}, ShuffleNet~\cite{shufflenet}, ShuffleNetV2~\cite{shufflenetv2} and Xception~\cite{xception}. These networks all expand the receptive field by pooling. The size of their output is small. However, high resolution masks contribute to improving the control accuracy of surgical robot. Therefore, the encoder-decoder architecture is adopted to obtain high-resolution masks. The decoder is bound to increase parameters and computational costs. To solve this issue, we design a lightweight decoder based on depth-wise separable convolution~\cite{mobilenet}. Depth-wise separable convolution can greatly reduce the amount of parameters and computational costs, which contributes to boosting the speed of the network. In this way, we can get the high-resolution mask in real time.

Different from common segmentation tasks, semantic segmentation of surgical instruments faces more challenges. To provide a good view, strong lighting conditions are required during the surgery, leading to severe specular reflections on surgical instruments. Specular reflection makes the surgical instrument white and changes its visual features such as color and texture. The network cannot identify surgical instruments by these changed features, making segmentation more difficult. Besides, shadows often appear in the field of view due to changes in illumination angle, movement of surgical instruments, and occlusion of human tissues. As shown in Fig.~\ref{show}, surgical instruments and background tend to darken in shadows. This issue not only changes the visual features of the surgical instrument but also makes it difficult to distinguish between surgical instruments and background. Also, sometimes only a part of the surgical instrument appears in the image due to movements and views, causing serious class imbalance. These issues make localization and semantic segmentation for surgical instruments more challenging.
%Also, sometimes only a part of the surgical instrument appears in the image due to movement, view, et., causing serious class imbalance.
%Class imbalance will reduce the segmentation accuracy.  To address these problems, attention is introduced to improve the feature representation of the network.
%In addition, Since the surgical instrument is constantly moving during the operation, its scale in the field of view changes with the movement. It is very difficult to correctly identify the surgical instrument.
%Another important issue is class imbalance issue.

To address these issues, an attention-guided lightweight network (LWANet) is proposed to segment surgical instruments in real-time. It adopts encoder-decoder architecture to get high-resolution masks, which can provide more detailed location information for robot control. A lightweight network, MobileNetV2~\cite{mobilenetv2}, is adopted as the encoder. It owns fast inference speed and has powerful feature extraction capabilities. Besides, we design a lightweight attention decoder to recover the location details. Depthwise separable convolution~\cite{mobilenet} is used as a basic unit to construct the decoder. It factorizes a standard convolution into two parts to reduce the computational costs and model size. To better recover location details, transposed convolution is used to perform upsampling in the decoder.
%In this way, our network can quickly and accurately segment robotic surgical instruments. Besides, Focal loss is used to train our network for addressing the imbalance problem.

Attention fusion block is designed to fuse high-level and low-level features. It introduces global average pooling to capture global contexts and encodes semantic dependencies between channels. Since different channels correspond to the various semantic response, this block can distinguish target regions and background by semantic dependencies between channels. By emphasizing the specific channels, it can focus on target regions and accurately locate surgical instruments, contributing to solving the specular reflection and shadow issues as well as improving the segmentation accuracy. Furthermore, attention fusion block only takes little computational costs, contributing to improving inference speed.
%It squeezes global context into an attentive vector which encodes the semantic dependencies between channels.

The contributions of our work are as follows:

1) An attention-guided lightweight network is proposed to segment surgical instruments in real-time. It has a small model size and takes little computational costs. The inference speed can reach 39fps with only 3.39 GFLOPs on 960$\times$544 inputs. Thus, it can be applied to real-time control of the surgical robot and real-time computer-assisted surgery.

2) Attention fusion block is designed to model semantic dependencies between channels and emphasize the target regions, which contributes to localization and semantic
segmentation for surgical instruments.

3) The proposed network achieves state-of-the-art performance 94.10\% mean IOU on Cata7 and obtains a new record on EndoVis 2017 with a 4.10\% increase on mean IOU.

\section{Related Work}

\subsection{Semantic Segmentation of Surgical Instruments}
In previous work, various methods have been proposed to segment surgical instruments~\cite{rasnet,raunet}. The Hybrid RNN-CNN method introduced the recurrent neural network in Full Convolutional Network (FCN) to capture global contexts, contributing to expanding the receptive field of convolution operations~\cite{attia}. RASNet~\cite{rasnet} adopted an attention mechanism to emphasize the target region and improve the feature representation. Qin \emph{et al}.~\cite{qin} fused the convolutional neural network predictions and the kinematic pose information to improve segmentation accuracy. Luis \emph{et al}.~\cite{Luis} presented a network based on FCN and optic flow to solve problems such as occlusion and deformation of surgical instruments. Another work~\cite{deep2017} used the residual network with dilated convolutions to segment surgical instruments. However, most of these work mainly focused on the improvement of segmentation accuracy while failed to segment surgical instruments in real-time.
%From mentioned above, it can be seen that convolutional neural network has achieved excellent performance in segmentation of surgical instruments.
%Iro \emph{et al}.~\cite{csl} proposed a novel U-shape network to provide segmentation and pose estimation of instruments simultaneously. It estimated the pose by regressing the heatmap.

\subsection{Light-Weight Network}
Due to the limitations of computing resources, the application of deep learning models in robot control remains a challenge. To make the neural network easier to apply, a series of lightweight networks is proposed. Light-Weight RefineNet modifies~\cite{lwrefinenet} the decoder of RefineNet~\cite{refinenet} to reduce the number of parameters and floating-point operations. MobileNet~\cite{mobilenet} introduced depthwise separable convolution instead of the traditional convolution to reduce the model size and computational costs. MobileNetV2~\cite{mobilenetv2} proposed the inverted residual structure to improve the ability of a gradient to propagate and save memory. The network used in~\cite{mobile_refinenet} consisted of mobilenetv2 and the decoder of Light-Weight RefineNet, which is used for semantic segmentation. Besides, there are lightweight networks applied in other tasks such as Shufflenet~\cite{shufflenet}, ShuffleNetV2~\cite{shufflenetv2}, SqueezeNet~\cite{squeezenet} and Xception~\cite{xception}. They are fast and memory-efficient.

\subsection{Attention Module}
In recent years, attention mechanisms have been widely used in the field of computer vision~\cite{pan,dan}. It can help the network focus on key regions by mimicking human attention mechanisms. Squeeze-and-excitation block~\cite{senet} squeezed global context into a vector to model the semantic dependencies between channels. Non-local block~\cite{nonlocal} extracted the global context to expand the receptive field. Dual Attention Network~\cite{dan} consisted of channel attention module and position attention module, modeling the semantic dependencies between positions and channels. These attention modules can be flexibly inserted into FCNs to improve their feature representation.

%Wang \emph{et al} introduced a position attention module called non-local block, which extracts the global context to expand the receptive field~\cite{nonlocal}. Fu \emph{et al}. designed a module that combines channel attention and positional attention~\cite{dan}.

%It capture the global context and model semantic dependencies to improve the feature representation.
\section{Methodology}
\begin{figure*}[tbp]
  \centering
  \includegraphics[width=0.98\textwidth]{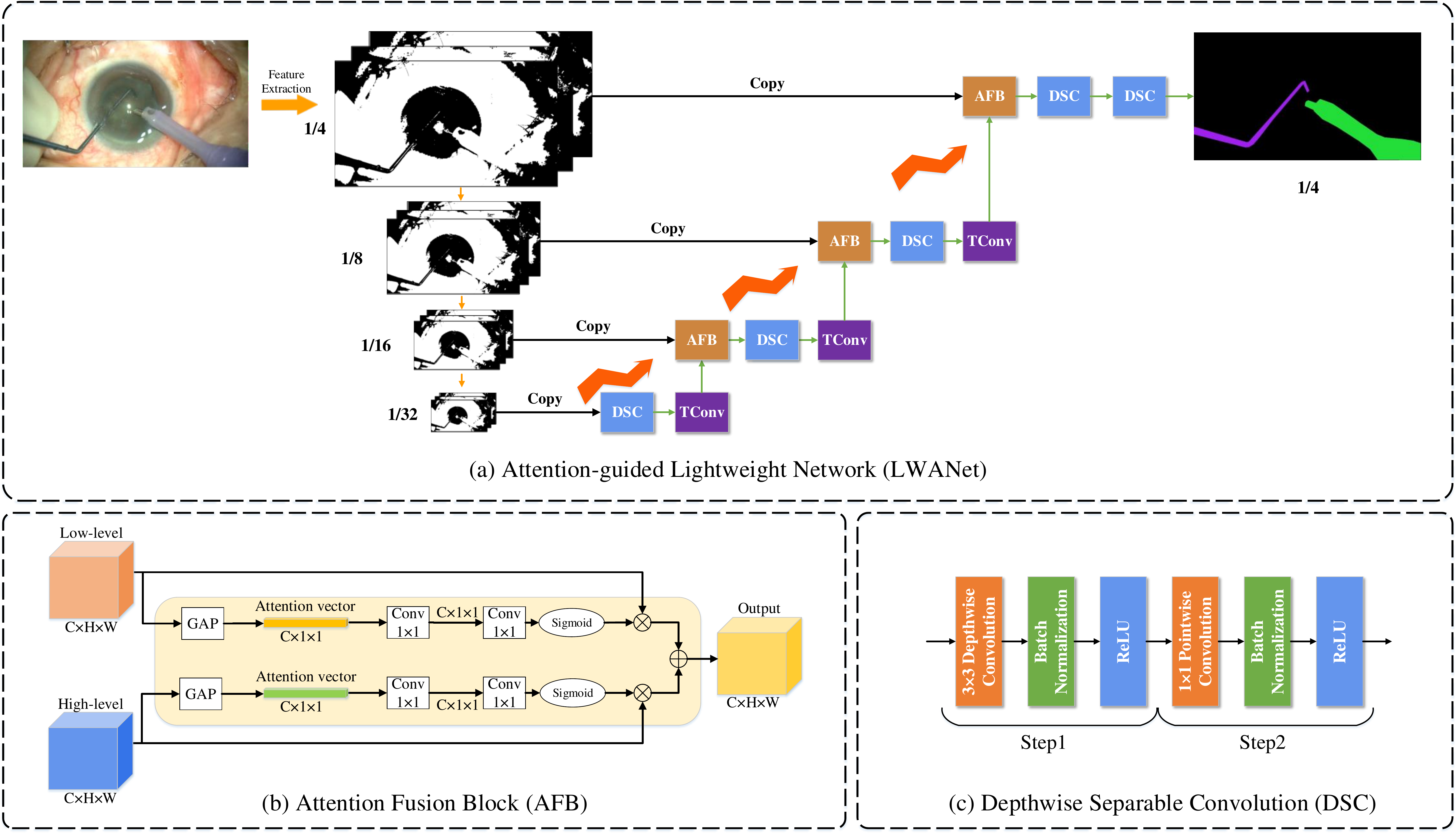}
  \caption{The architecture of Attention-guided Lightweight Network and its components. (a) Attention-guided Lightweight Network: it adopts the encoder-decoder architecture. (b) Attention Fusion Block (c) Depthwise Separable Convolution}
  \label{lwanet}
\end{figure*}
\subsection{Overview}
Due to the limitation of computing resources, the application of deep learning models in robots is very difficult. To address this issue, we propose the attention-guided lightweight network (LWANet) to segment robotic instruments in real-time. It adopts encoder-decoder architecture to acquire high-resolution masks and provide detailed location information. The architecture of LWANet is shown in Fig.~\ref{lwanet}. To reduce computational costs, a lightweight network, MobileNetV2, is used as an encoder to extract semantic features. It is based on the inverted residual block, which is fast and memory efficient. The last two layers of mobilenetv2 are dropped, including the average pooling layer and the fully connected layer. They are not suitable for semantic segmentation task. The output scale of the MobilenetV2 is 1/32 of the original image. Upsampling is bound to increase the computational cost of the network. Therefore, a lightweight attention decoder is designed to recover position details. It only takes little computational costs, contributing to real-time segmentation for surgical instruments. The output scale of LWANet is 1/4 of the original image. The lightweight attention decoder will be introduced in detail next.
%, which  consists of Depthwise separable convolution, transposed convolution and attention fusion block

\subsection{Lightweight Attention Decoder}
The lightweight attention decoder consists of depthwise separable convolution~\cite{mobilenet}, attention fusion block, and transposed convolution. The depthwise separable convolution is used as the basic unit of the decoder, contributing to reducing computational costs. Attention fusion block captures global contexts and encodes semantic dependencies between channels to focus on target regions. Besides, transposed convolution is adopted to perform upsampling.

\subsubsection{Depthwise Separable Convolution}
Depthwise separable convolution is adopted as the basic unit of the decoder, replacing the standard convolution. Depthwise separable convolution factorizes a standard convolution into a depthwise convolution and a pointwise convolution, breaking the interaction between the size of the kernel and the channels of output~\cite{mobilenet}. In this way, it can reduce the computational cost. Its architecture is shown in Fig.~\ref{lwanet}(c). We consider a case that a convolution takes a $d1\times m\times n$ feature map as input and produces a $d2\times m\times n$ feature map, where $d1$ and $d2$ is the number of feature map channels. When the kernel size is $k\times k$, the computational cost of standard convolution is $k\times k\times d1\times d2\times m\times n$. The computational cost of depthwise separable convolution is $k\times k\times d1\times m\times n+d1\times d2\times m\times n$~\cite{mobilenet}.
\begin{equation}
\frac{{k \times k \times d1 \times m \times n + d1 \times d2 \times m \times n}}{{k \times k \times d1 \times d2 \times m \times n}} = \frac{1}{{d2}} + \frac{1}{{{k^2}}}
\end{equation}

By using the depthwise separable convolution, the computational cost is reduced by $\frac{1}{{d2}} + \frac{1}{{{k^2}}}$ times~\cite{mobilenet}.
Usually, $d2$ is so large that $1/d2$ can be ignored. When the kernel size is 3$\times$3, the computational cost is reduced by about 9 times.

%It is proposed to replace standard convolution in~\cite{mobilenet}, which Thus, depthwise separable convolution can greatly reduce the computational cost.
\subsubsection{Attention Fusion Block}
Attention fusion block(AFB) is introduced to fuse the high-level feature map and low-level feature map. Since different channels correspond to various semantic responses, a channel attention mechanism called squeeze-and-excitation mechanism~\cite{senet} is introduced to encode semantic dependencies between channels. This attention mechanism is performed on low-level and high-level features separately to extract different-level attentive features, which is shown in Fig.~\ref{lwanet}(b). In this way, we can not only emphasize target location details in low-level feature maps but also capture the global context and semantic information in high-level feature maps to improve feature representation.
%Low-level feature maps contain rich location details and lack semantic features.
%In addition, the global context in high-level feature map contributes to expanding the receptive field.
%An attention module is introduced by us.

Global average pooling is essential to capture global contexts and encode semantic dependencies~\cite{pan,dan}. It squeezes global contexts into an attentive vector to encode semantic dependencies between channels. Then, the attentive vector is transformed by convolutions to further capture semantic dependencies. The generation of the attentive vector is shown in Eq.(2). The output $\widehat x$ is generated by Eq.(4).
\begin{equation}
{A_c} = {\delta _2}\left[ {{W_\beta } \cdot {\delta _1}\left[ {{W_\alpha } \cdot g(x) + {b_\alpha }} \right] + {b_\beta }} \right]
%{Z_a} = Sigm\left[ {{W_\beta } \cdot {ReLU}\left[ {{W_\alpha } \cdot g(x) + {b_\alpha }} \right] + {b_\beta }} \right]
\end{equation}
where $x$ refers to input feature map. $g$ refers to the global average pooling. ${\delta _1}$ refers to ReLU function and ${\delta _2}$ refers to Sigmoid function. ${W_\alpha },{W_\beta }$ are parameters of 1$\times$1 convolution. ${b_\alpha },{b_\beta }$ are biases.
\begin{equation}
g({x_k}) = \frac{1}{{w \times h}}\sum\limits_{i = 1}^h {\sum\limits_{j = 1}^w {{x_k}(i,j)} }
\end{equation}
where $k = 1,2,...,d$ and ${x} = \left[ {{x_1},{x_2},...,{x_d}} \right]$.
\begin{equation}
\widehat x = {A_c} \otimes x
\end{equation}
where $\otimes$ denotes broadcast element-wise multiplication.

Finally, two attentive feature maps are merged by addition. Addition can reduce parameters of convolution compared with concatenation, contributing to reducing computational costs.
%The attention fusion block performs this attention mechanism on both the low-level and the high-level feature map simultaneously,
\subsubsection{Transposed Convolution}
The decoder recovers the position details and obtains high-resolution feature maps by upsampling. However, upsampling often results in blurred edges and reduces image quality. To address this issue, transposed convolution is introduced to perform upsampling.
It can learn the weights to suit various objects, helping preserve edge information. In this way, we can acquire refined edges and improve segmentation accuracy.
%Its parameters are learnable.
\subsection{Transfer Learning}
Surgical videos or images are difficult to obtain. Also, the annotation for the surgical instrument takes a lot of time and costs. Thus, a transfer learning strategy is adopted to overcome this difficulty. We use samples from other tasks to improve the segmentation accuracy for surgical instruments. In our network, the encoder MobileNetV2~\cite{mobilenet} is pre-trained on the ImageNet. Images in the ImageNet are all from life scenes. By pre-training, the network can learn low-level features such as boundary, color, and texture of objects. These features can also be applied in surgical scenes. In this way, the encoder has a better ability to extract low-level features. Then the network is trained on surgical instrument datasets to capture high-level semantic features of instruments. This strategy improves network performance and accelerates network convergence.
%reduce the cost of sample annotation.

\subsection{Loss Function}
The class imbalance issue is more severe in the surgical instrument segmentation task than other common segmentation tasks.
To address this issue, we adopt focal loss~\cite{focalloss} to train our network. It reduces the weight of easy samples, making the model more focused on hard samples during training.
Focal loss is shown in Eq. (5).
\begin{equation}
FL({p_t}) =  -{(1 - {p_t})^\gamma }\log ({p_t})
%F =  - \sum\limits_{i = 1}^w {\sum\limits_{j = 1}^h {\sum\limits_{k = 1}^c {{y_{ijk}}{{(1 - {p_t})}^\gamma }\log ({p_t})} } } ,{p_t} = \frac{{{e^{{{\widehat y}_{ijk}}}}}}{{\sum\limits_{k = 1}^c {{e^{{{\widehat y}_{ijk}}}}} }}
%F=- \sum\limits_{i = 1}^w {\sum\limits_{j = 1}^h {\sum\limits_{k = 1}^d {{y_{ijk}}{{(1 - \frac{{{e^{{{\widehat y}_{ijk}}}}}}{{\sum\limits_{k = 1}^d {{e^{{{\widehat y}_{ijk}}}}} }})}^\gamma }\log (\frac{{{e^{{{\widehat y}_{ijk}}}}}}{{\sum\limits_{k = 1}^d {{e^{{{\widehat y}_{ijk}}}}} }})} } }
\end{equation}
%where $y$ is the ground truth and $\widehat y$ is the prediction.
where $\gamma$ is used to adjust the weight of examples. $\gamma\ge0$.

%To perform delicate operations, surgical instruments tend to be small or thin. Therefore, surgical instruments may only occupy a small region in the image. The number of foreground pixels is much less than that of backgrounds, which causes serious class imbalance. Semantic segmentation can be seen as classifying each pixel. Class imbalance issue will greatly reduce the segmentation accuracy.

\begin{table*}[tbp]
  \centering
  \caption{Performance comparison of a series of networks}
  \setlength{\tabcolsep}{2.5mm}{
    \begin{tabular}{c|c|c|c|c|c|c|c|c}
    \hline
    \hline
    \textbf{Method} & \textbf{Encoder} & \textbf{Decoder} & \textbf{mDice(\%)} & \textbf{mIOU(\%)} & \textbf{Parameters} & \multicolumn{1}{c|}{\textbf{GFLOPs}} & \textbf{Time(ms)} & \textbf{FPS}\\
    \hline
    U-Net~\cite{unet} & - & - &   86.83  &   78.21  &   7.85M    &   106.18    & 50.00& 20.00\\
    TernausNet~\cite{ternausnet} & VGG11 & - &   96.24   &  92.98 &    25.36M   &   219.01    & 78.92 &12.67\\
    LinkNet~\cite{linknet} & ResNet50 & - &  95.62  & 91.86 &  31.27M   &  74.45  & 44.50& 22.47\\
    LW-RefineNet~\cite{lwrefinenet} & ResNet50 & LW-Refine &    96.16   &  92.74  &   27.33M    &   63.34    & 46.89 & 21.33\\
    MobileV2-RefineNet~\cite{mobile_refinenet} & MobileNetV2 & LW-Refine &   96.33    &  93.07  &  3.01M     &    16.62   & 39.63 &25.23\\
    \hline
    %LWDNet-MobileV1 & MobileNetV1 & attention-lw &       &       &       &       &  \\
    % LWDNet-ShuffleV2 & ShuffleNetv2 & attention-lw &       &       &       &       &  \\
    LWANet without AFB & MobileNetV2 & LW-Decoder &    95.80   &    92.18 &  2.03M    &   3.38    &  -&-\\
    LWANet(Ours) & MobileNetV2 & LWA-Decoder &    96.91   &    94.10   &   2.06M    &   3.39    & 25.32 &39.49\\
    \hline
    \hline
    \end{tabular}%
    }
  \label{compare}%
\end{table*}%
\section{Experiments and Results}
The proposed LWANet is evaluated on two datasets, including Cata7~\cite{raunet} and EndoVis 2017~\cite{endovis2017} datasets.
%In experiments, the inference speed of LWANet can reach 39fps with only 3.39 GFLOPs on 960$\times$544 inputs. Besides, it achieves best performance 94.10$\%$ mean IOU on Cata7 and obtains a new record on EndoVis 2017 with 4.10\% increase on mIOU.
\subsection{Dataset}
%In experiments, two datasets are used, including Cata7 Dataset and EndoVis 2017 Dataset. These two datasets are based on cataract surgery and endoscopic surgery, respectively.

Cata7 dataset is a cataract surgical instrument dataset for the semantic segmentation of surgical instruments, which is constructed by us. It contains 2500 frames with a resolution of 1920$\times$1080, consisting of 1800 frames for training and 700 frames for test. These images are split from 7 cataract surgery videos at 30fps. The images in the training set and the test set are from different video sequences. There are 10 types of surgical instruments in cata7.

EndoVis 2017 dataset is from the MICCAI Endovis Challenge 2017. This dataset is based on endoscopic surgery, acquired by a Vinci Xi robot. It contains 3000 images with a resolution of 1280$\times$1024, which contains 1800 images for training and 1200 images for test. There are 7 types of surgical instruments in EndoVis 2017.
\subsection{Implementation}
Our network is implemented in PyTorch. All experiments are performed on an Nvidia Titan X which has 12 G memory. Adam is used as an optimizer, which takes default parameters of PyTorch. The batch size is 16 in training. To prevent overfitting, we use a strategy to adjust the learning rate. For every 30 iterations, the learning rate is multiplied by 0.8. After a series of experiments, the parameter $\gamma$ of focal loss is set to 6. All the networks are trained based on the above strategies. Only the initial learning rate is different. The Dice coefficient and Intersection-Over-Union(IOU) are selected as evaluation metrics.

Data augmentation is performed to increase the diversity of samples, contributing to improving network performance. The augmented samples are generated by random rotation, shifting, and flipping.

\subsection{Cata7}
To verify the excellent performance of the network, a series of experiments are performed based on Cata7. The images in Cata7 are resized to 960$\times$544 due to the limitations of computing resources. The initial learning rate is 0.0002. 800 images are generated by data augmentation. All experimental results are shown in Table~\ref{compare}. The inference time is calculated including data transfer from CPU to GPU and back and averaged across 667 inferences.

As shown in Table~\ref{compare}, our network achieves state-of-the-art performance 96.91\% mean Dice and 94.10\% mean IOU. Among other methods, MobileV2-RefineNet~\cite{mobile_refinenet} achieves the best performance. Compared with it, the mean Dice and mean IOU are increased by 0.58\% and by 1.03\%, respectively. Besides, the encoder of MobileV2-RefineNet~\cite{mobile_refinenet} is the same as our network while the decoder is different. This indicates that the proposed lightweight attention decoder (LWA-Decoder) has excellent performance.
%MobileV2-RefineNet~\cite{mobile_refinenet} uses mobilenetv2 as the encoder and uses the decoder of lightweight refinenet~\cite{lwrefinenet}.
%This result shows that the LWA-Decoder designed by us has better performance.

The model size of LWANet is small. It only has 2.06M parameters. Lightweight RefineNet~\cite{lwrefinenet} and MobileV2-RefineNet~\cite{mobile_refinenet} are existing state-of-the-art lightweight networks for semantic segmentation. The model size of them is 27.33M and 3.01M, respectively. Compared with MobileV2-RefineNet~\cite{mobile_refinenet}, the model size of LWANet is reduced by approximately 31.56\%. Also, the model size of lightweight Refinenet~\cite{lwrefinenet} is 13.27 times that of LWANet.

Furthermore, LWANet can segment surgical instruments in real-time. As shown in Table~\ref{compare}, our LWANet can process an image within 26ms. The inference speed is approximately 39 fps. The frame rate of the original surgical video is 30 fps which is much lower than the inference speed of LWANet. Therefore, the network can segment surgical instruments in real-time based on 960$\times$544 inputs. Under the same conditions, the inference speed of MobileV2-RefineNet~\cite{mobile_refinenet} is approximately 25 fps. Meanwhile, the inference speed of Lightweight RefineNet~\cite{lwrefinenet} is approximately 21 fps. In contrast, the inference speed of LWANet has increased by 14 fps and 18 fps, respectively.
\begin{table}[tbp]
  \centering
  \caption{Comparison of the computational cost between LWANet and other methods}
  \setlength{\tabcolsep}{0.8mm}{
    \begin{tabular}{c|c|c|c|c|c}
    \hline
    \hline
    \multirow{2}{*}{\textbf{Method}} & \multirow{2}{*}{\textbf{GFLOPs}} & \multicolumn{2}{c|}{\textbf{Encoder}} & \multicolumn{2}{c}{\textbf{Decoder}} \\
\cline{3-6}          &       & \textbf{GFLOPs} & \multicolumn{1}{c|}{\textbf{Percen.}} & \textbf{GFLOPs} & \multicolumn{1}{c}{\textbf{Percen.}} \\
    \hline
    U-Net~\cite{unet} & 106.18 & 28.85 & 27.17\% & 77.33 & 72.83\% \\
    TernausNet~\cite{ternausnet} & 219.01 & 81.42 & 37.18\% & 137.59 & 62.82\% \\
    LinkNet~\cite{linknet} & 74.45 & 42.88 & 57.60\% & 31.57 & 42.40\% \\
    LW-RefineNet~\cite{lwrefinenet} & 63.34 & 42.88 & 67.70\% & 20.46  & 32.30\% \\
    Mobile-RefineNet~\cite{mobile_refinenet} & 16.62  & 3.11  & 18.71\% & 13.51  & 81.29\% \\
    \hline
    LWANet(Ours) & 3.39  & 3.11  & 91.74\% & 0.28  & 8.26\% \\
    \hline
    \hline
    \end{tabular}%
    }
  \label{gflops}%
\end{table}%

\begin{table*}[tbp]
  \centering
  \caption{Segmentation results on Endovis 2017 dataset.NCT, UB and UA are the university abbreviation of the participating team~\cite{endovis2017}.}
    \setlength{\tabcolsep}{2mm}{
    \begin{tabular}{c|c|c|c|c|c|c|c|c|c|c|c}
    \hline
    \hline
     & Dataset 1 & Dataset 2 & Dataset 3 & Dataset 4 & Dataset 5 & Dataset 6 & Dataset 7 & Dataset 8 & Dataset 9 & Dataset 10 & mIOU \\
    \hline
    TernausNet & \textbf{0.177} & \textbf{0.766} & 0.611 & 0.871 & 0.649 & 0.593 & 0.305 & 0.833 & \textbf{0.357} & 0.609 & 0.542 \\
    ToolNet & 0.073 & 0.481 & 0.496 & 0.204 & 0.301 & 0.246 & 0.071 & 0.109 & 0.272 & 0.583 & 0.337 \\
    SegNet & 0.138 & 0.013 & 0.537 & 0.223 & 0.017 & 0.462 & 0.102 & 0.028 & 0.315 & \textbf{0.791} & 0.371 \\
    NCT   & 0.056 & 0.499 & \textbf{0.926} & 0.551 & 0.442 & 0.109 & 0.393 & 0.441 & 0.247 & 0.552 & 0.409 \\
    UB   & 0.111 & 0.722 & 0.864 & 0.68  & 0.443 & 0.371 & 0.416 & 0.384 & 0.106 & 0.709 & 0.453 \\
    UA    & 0.068 & 0.244 & 0.765 & 0.677 & 0.001 & 0.400   & 0.000  & 0.357 & 0.040  & 0.715 & 0.346 \\
    \hline
    Ours& 0.096 & 0.758 & 0.889 & \textbf{0.898} & \textbf{0.761} & \textbf{0.627} & \textbf{0.454} & \textbf{0.875} & 0.230 & 0.763 & \textbf{0.583} \\
     %\hline
%    mIOU(\%) & 0.104 & 0.506 & 0.731 & 0.593 & 0.384 & 0.400 & 0.246 & 0.431 & 0.236 & 0.688 & 0.447 \\
    \hline
    \hline
    \end{tabular}%
    }
  \label{endovis2017}%
\end{table*}%

We also evaluate the computational cost of LWANet. Floating-point operations per second (FLOPs) is used as the evaluation metric. As shown in Table~\ref{gflops}, the FLOPs of LWANet is 3.39G, of which encoder accounts for a large proportion of 91.74\%. The FLOPs of the decoder only accounts for 8.26\% of the total.
The FLOPs of MobileV2-RefineNet~\cite{mobile_refinenet} is 16.62G, which is 4.9 times of LWANet. Besides, the FLOPs of its encoder only accounts for 18.71\% while the decoder accounts for 81.29\% of the total. The encoders of these two networks are the same. Thus, it can be found that the lightweight attention decoder designed by us has lower computational costs and better performance. Also, the FLOPs of Light-Weight RefineNet~\cite{lwrefinenet} is 63.34G, which is 18.68 times of LWANet. These results show that the computational cost of LWANet and LWA-Decoder are low.
%We separately count the FLOPs of the encoder and decoder. The FLOPs of encoder is much small than that of decoder.

Attention fusion block(AFB) is adopted to help the network focus on key regions. Ablation experiments for AFB are performed to verify its performance. The results are shown in Table~\ref{compare}. LWANet without AFB achieves 95.80\% mean Dice and 92.18\% mean IOU. LWANet with AFB achieves 96.91\% mean Dice and 94.10\% mean IOU. Via employing AFB, mean Dice has increased by 1.11\% and mean IOU has increased by 1.92\%. These results show that AFB contributes to improving segmentation accuracy.
\begin{figure}[tbp]
\centering
\includegraphics[width=0.48\textwidth]{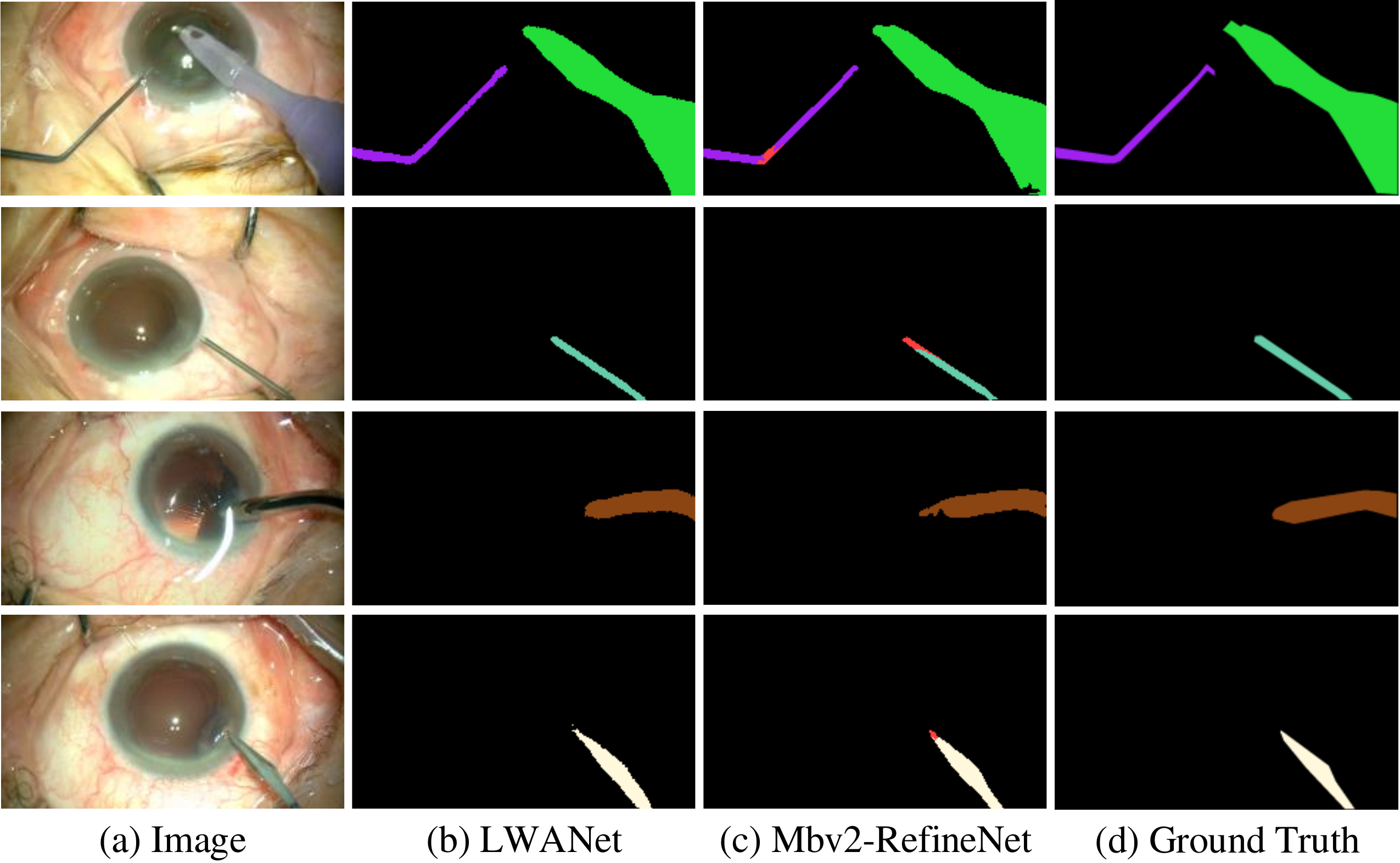}
\caption{Visualization results of LWANet on Cata7. Different types of surgical instruments are marked by different colors.}
\label{mask}
\end{figure}

% Table generated by Excel2LaTeX from sheet 'Sheet2'
\begin{table}[tbp]
  \centering
  \caption{Ablation experiments for transfer learning}
  \setlength{\tabcolsep}{4mm}{
    \begin{tabular}{c|c|c|c}
    \hline
    \hline
    \textbf{Method} & \textbf{Pre-trained} & \textbf{mDice(\%)} & \textbf{mIOU(\%)} \\
    \hline
    LWANet & No    & 91.64 & 86.20 \\
    LWANet & Yes   & 96.91 & 94.10 \\
    \hline
    \hline
    \end{tabular}%
    }
  \label{transfer}%
\end{table}%

Transfer learning strategy can improve network performance and accelerate network convergence. Some experiments are set to verify the validity of this strategy, which is shown in Table~\ref{transfer}. The network without pre-training has only achieved 91.64\% mean Dice and 86.20\% mean IOU. By employing the transfer learning strategy, mean Dice has increased by 5.27\% and mean IOU has increased by 7.90\%.

To give a more intuitive result, we visualize the segmentation results. The visualization results are shown in Fig.~\ref{mask}. There are misclassifications in the results of MobileV2-RefineNet~\cite{mobile_refinenet}. Meanwhile, the results of LWANet are the same as the ground truth, which is because attention fusion block helps our network focus on key regions. Also, due to the use of focal loss, our network can effectively solve the class imbalance problem.
\subsection{EndoVis 2017}
LWANet is also evaluated on the Endovis 2017 dataset~\cite{endovis2017}. The images in EndoVis 2017 is resized to 640$\times$544 due to the limitation of computing resources. The initial learning rate is 0.0002. The batch size is 16. The test set consists of 10 video sequences. Each sequence contains specific surgical instruments. The test performance results are reported in Table ~\ref{endovis2017}. TernausNet~\cite{ternausnet}, ToolNet~\cite{toolnet} and SegNet~\cite{segnet} are evaluated on EndoVis2017. The test results of other methods are from the MICCAI EndoVis challenge 2017~\cite{endovis2017}.

LWANet achieves 58.30\% mean IOU, which outperforms other methods. It achieves the best results in 5 video sequences and takes the second place in 3 video sequences. The best of the existing methods is TernausNet~\cite{ternausnet}. TernausNet~\cite{ternausnet} achieves 54.20\% mean IOU and achieves the best results in 3 video sequences. Compared with it, the performance of our network improves by 4.10\% mean IOU.

\begin{table}[tbp]
  \centering
  \caption{Comparison of inference speed and computational cost}
  \setlength{\tabcolsep}{3mm}{
    \begin{tabular}{c|c|c|c|c}
    \hline
    \hline
    \textbf{Input Size} & \textbf{Method} & \textbf{GFLOPs} & \textbf{Tims(ms)} & \textbf{FPS} \\
    \hline
    640$\times$512 & LWANet & 2.12  & 23.90 & 41.85\\
    448$\times$352 & LWANet & 1.02  & 21.97 & 45.52 \\
    320$\times$56 & LWANet & 0.53  & 18.88 & 52.97\\
    \hline
    \hline
    \end{tabular}%
    }
  \label{speed}%
\end{table}%

Our network can segment surgical instruments in real-time. Comparison of inference speed and computational costs based on different input sizes is shown in Table~\ref{speed}.  The inference time is calculated including data transfer from CPU to GPU and back. It averaged across 600 inferences. The inference speed of LWANet can reach about 42 fps when the size of the input image is 640*512, which is much faster than the frame rate of surgical videos. As the input size decreases, the inference speed increases and the computational cost decreases.

\begin{figure}[tbp]
  \centering
  \includegraphics[width=0.48\textwidth]{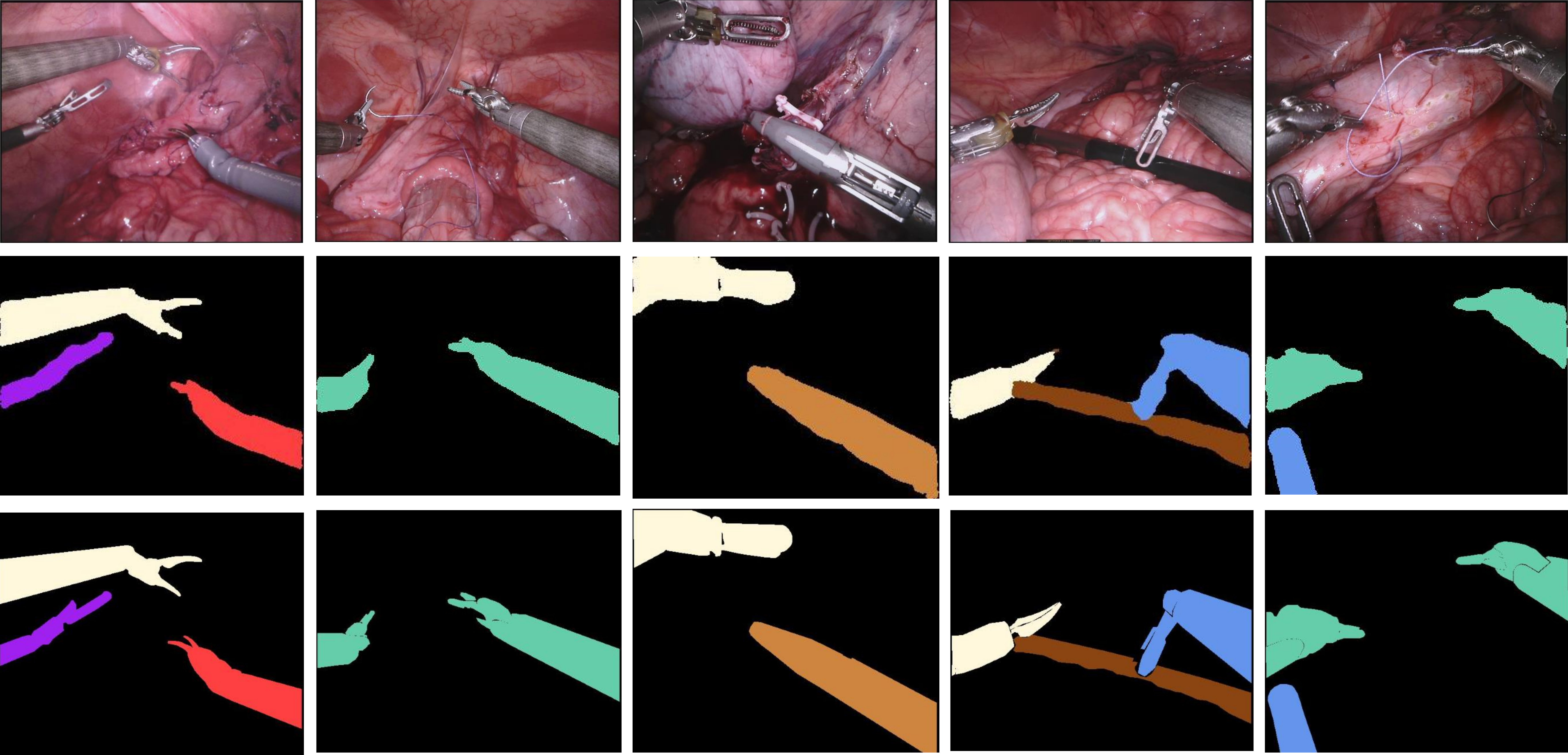}
  \caption{Visualization results of LWANet on EndoVis 2017. From top to bottom: image, prediction and ground truth. Different types of surgical instruments are marked by different colors.}
  \label{endo_show}
\end{figure}

To give a more intuitive result, the segmentation results are visualized in Fig.~\ref{endo_show}. Despite problems such as specular reflections and shadows, our network still can segment surgical instruments well. The results above prove that our network achieves state-of-the-art performance.

%\begin{table}[htbp]
%  \centering
%  \caption{Add caption}
%   \setlength{\tabcolsep}{3mm}{
%    \begin{tabular}{c|c|c|c|c}
%    \hline
%    \hline
%    Input size & Method & GFLOPs & Tims(ms) & FPS \\
%    \hline
%    960*544 & LWANet & 3.39  & 25.32 & 39.48 \\
%    640*384 & LWANet & 1.59  & 21.91 & 45.62 \\
%    480*288 & LWANet & 0.9   & 17.87 & 55.95 \\
%    \hline
%    \hline
%    \end{tabular}%
%    }
%  \label{speed}%
%\end{table}%

\section{CONCLUSIONS}
In this paper, we propose an attention-guided lightweight network named LWANet for real-time segmentation of surgical instruments. It can segment surgical instruments in a real-time while takes very low computational costs. Besides, experiments prove that our network achieves state-of-the-art performance on Cata7 and EndoVis 2017 datasets. This model can be used for surgical robot control and computer-assisted surgery, which is significant for clinical work.

\section{Acknowledgments}
This research is supported by the National Key Research and Development Program of China (Grant 2017YFB1302704), the National Natural Science Foundation of China (Grants 61533016, U1713220), the Beijing Science and Technology Plan(Grant Z191100002019013) and the Youth Innovation Promotion Association of the Chinese Academy of Sciences (Grant 2018165).

\bibliographystyle{IEEEtran}
\bibliography{icra}

\end{document}